\documentclass[a4paper]{article}

\usepackage{bm}
\usepackage[hidelinks]{hyperref}
\usepackage{xcolor}
\newcommand{\ie}{\textit{i.e.},}
\newcommand{\eg}{\textit{e.g.},}

\usepackage{tipa}

\usepackage{graphicx}
\usepackage{tikz}
\usetikzlibrary{decorations.text}

\usepackage{spconf,amsmath,graphicx}
\ninept

\usepackage{hyperref}

\newcommand\blfootnote[1]{%
  \begingroup
  \renewcommand\thefootnote{}\footnote{#1}%
  \addtocounter{footnote}{-1}%
  \endgroup
}

\title{Measuring the Impact of Domain Factors in Self-supervised Pre-Training}
\name{Ramon Sanabria$^{*}$, Wei-Ning Hsu$^{\bigtriangleup}$, Alexei Baevski$^{\bigtriangleup}$, Michael Auli$^{\bigtriangleup}$}
\address{$^{*}$The University of Edinburgh\\
$^{\bigtriangleup}$~Meta AI\hspace{0.2in}}

\begin{document}
\maketitle
\begin{abstract}
Human speech data comprises a rich set of domain factors such as accent, syntactic and semantic variety, or acoustic environment.
Previous work explores the effect of domain mismatch in automatic speech recognition between pre-training and fine-tuning as a whole~\cite{hsu2021robust} but does not dissect the contribution of individual factors.
In this paper, we present a controlled study to better understand the effect of such factors on the performance of pre-trained representations on automatic speech recognition. 
To do so, we pre-train models either on modified natural speech or synthesized audio, with a single domain factor modified, and then measure performance after fine-tuning.
Results show that phonetic domain factors play an important role during pre-training while grammatical and syntactic factors are far less important. 
To our knowledge, this is the first study to better understand the domain characteristics of pre-trained sets in self-supervised pre-training for speech.
\end{abstract}

\section{Introduction}
\blfootnote{$^*$Work done during an internship at Meta AI.}Self-supervised learning has gained considerable traction in the speech community and showed consistent improvements in automatic speech recognition (ASR) tasks~\cite{chung2019unsupervised, schneider2019wav2vec, baevski2020wav2vec, conneau2020unsupervised}, as well as other speech-related tasks~\cite{yang2021superb,polyak2021speech,lee2021textless, lee2021direct}. Concretely, by pre-training a model with large amounts of unlabeled data, and then fine-tuning it on a downstream task (\eg\ ASR, sentiment analysis, etc.), previous work rise the state-of-the-art by orders of magnitude.

Previous work on this topic relies on public datasets that often cover only a rather narrow domain. Consequently, recent work studied the effects of domain shift during pre-training~\cite{hsu2021robust}. Concretely, in this work, authors study the impact of domain mismatch between the pre-training and fine-tuning sets by using data from different domains such as audiobooks \cite{kahn2020libri} or conversational speech \cite{godfrey1992switchboard}. Among other things, the authors state that a mismatch in the domain between the pre-training and fine-tuning sets leads to a decrease in performance, as compared to a perfect domain match. 

However, in the context of ASR, the term "domain" typically encompasses multiple factors, including but not limited to vocabulary, accent (\eg\  prosody, syntax, and lexicon variation), and acoustics (\eg\ wide and narrowband). A sparse combination of each one of these factors results on the domain definition of each dataset. For that reason, in domain-mismatch studies such as \cite{hsu2021robust}, where authors combine different datasets, domain factors get interlinked and are hard to quantify the effect that each individual factor has.

To measure the contribution of each domain factor, we present a controlled experiment. Specifically, we propose to automatically generate the pre-training sets altering individual domain factors such as phonotactic, lexical, syntactic, and prosody variation as well as speaker variability. We apply this variation using audio manipulation and synthesis (\textsection\ref{sec:datasetcreation}). After pre-training the model on the controlled pre-training set, we fine-tune it with real (human-generated) speech without any modification, and measure the impact that each factor has. It should be noted that while our generative approach does introduce some artifacts into the speech signal, this is a necessary trade-off to control each individual domain dimension -- we hope that our findings are generalizable to natural data.

We perform experiments using wav2vec2.0~\cite{baevski2020wav2vec}, with the 960 and 10-hour sets from the Librispeech corpus for pre-training and fine-tuning, respectively.
Our experiments show that speaker variation in the pre-training data is crucial. Also, low-level domain dimensions such as phonotactics and prosody variation have a higher impact on final performance than syntactic and lexical variation. Concretely, while modifying phonotactics (without altering the phone set) hurts performance considerably, changing word order barely has an effect.
We also find that large amounts of synthesized data, with enough speaker variety, can perform considerably better than pre-training on 44 hours of natural speech -- which can be used to train a synthesizer in the first place. 
\section{Domain}
In speech, we define domain as \textit{a group of characteristics shared across a dataset composed by speech utterances}\footnote{We obviate timbre and other idiosyncratic speaker dimensions which can not be grouped.}. 
We consider the following factors:
\begin{itemize}
    \item \textbf{Acoustic Domain:} The acoustic environment where the utterance has been recorded (\ie\ microphone type, acoustic conditions, signal processing drivers).
    \item \textbf{Syntactic Domain:} The word order that utterances follow. In other words, this aspect can be modeled with the word-level n-gram probabilities -- transition probabilities between words.
    \item \textbf{Lexic/Semantic Domain:} The lexicon used for the speakers. For instance, some speakers may use technical vocabulary (\eg\ experts on biology), while other may have more informal conversation (\eg\ a day to day conversation).
    \item \textbf{Phonetic Domain:} In a set of speakers from the same languages, that use the same Lexic, Syntactic, and Acoustic domain, we can still find speakers with different phone sets, phonotactic patterns, prosody (\ie\ rhythm, pitch, and volume fluctuation across the utterance). 
\end{itemize}

In this work, for the sake of space we will focus on the latter three factors since, and leave the rest for future work. 

\section{Methodology}
Our pipeline has three steps. First, we create a pre-training set with an altered domain dimension by either generating one from scratch with a speech synthesizer or by modifying the original set. Second, we pre-train the self-supervised model with the altered set. Third, we fine-tune the model on a labeled unmodified set for final evaluation. The pre-training set is composed by 960h of data (natural or synthesized) -- unless specified otherwise -- and we fine-tune it on the 10-hour subset of Libri-light~\cite{kahn2020libri}. Because we do not need to compare to previous work, we only report results on the development sets from Librispeech -- dev-other/clean sets.

\subsection{Dataset Creation}
\label{sec:datasetcreation}
To create a controlled experimental setup, we either synthesize a modified phonetic sequence  (\textit{Synthesis}) or modify the original audio (\textit{Audio Modification}). Both processes alter \textit{only a single dimension} of the original domain of the data. In what follows, we describe each technique further.


To synthesize data, we use the publicly released weights from fairseq $S^2$~\cite{ott2019fairseq,wang-etal-2021-fairseq,f2url}. Concretely, we use their implementation of Fastspeech 2 \cite{ren2020fastspeech} with  HiFi GAN vocoder \cite{kong2020hifi}.

\textbf{VCTK:} We train this model on VCTK \cite{vctk}, which contains 44 hours of speech from 109 speakers. We will use it to measure the contribution of speaker variability during pre-training. Because VCTK is composed mostly for shorter utterances, the quality of the synthesizer trained on it degrades dramatically for long sentences. To solve this issue, we synthesize six words independently at a time and concatenate the resulting synthesized speech. This workaround produces speech of sufficient quality for our experiment setting.

\textbf{LJSpeech:} We train this model with the LJSpeech dataset \cite{ljspeech17}. LJSpeech is composed of 24 hours of speech from a single speaker. Because the utterances from this set are longer, synthesizers trained on this data model better longer dependencies and, consequently, prosody (\ie\ rhythm variations at the utterance level). LJSpeech only has data from one only speaker. The quality of the generated voice the quality of the speech generated is higher than VCTK's$^2$.

To contrast our synthesis speech results, we also alter domain factors by modifying the original audio signal. We do this by segmenting the audio utterance at the phone or word level, reorder each unit, and concatenate them back. We do not apply any post-processing effects on the concatenated region to not complicate further the audio modification pipeline. This method preserves the natural variations present in human speech. However, it introduces discontinuities in the signal that generates unnatural artifacts.  To detect the start and end of phone units, we use the Montreal Forced Aligner \cite{montreal}. We then detect word boundaries by using the start time of the first phone and the end time of the last one.

\subsection{Model, Training and Data}
\label{sec:traininganddata}

We use  wav2vec 2.0~\cite{baevski2020wav2vec,conneau2020unsupervised} which is composed of a feature extractor and a subsequent transformer model. 
The feature extractor $f : \mathcal{X} \rightarrow \mathcal{Z}$ maps raw audio $\mathcal{X}$ to hidden representations $\mathcal{Z}$. This hidden representation is used as input to the transformer model $g : \mathcal{Z} \rightarrow \mathcal{C}$ that outputs a context representation $\bm{c}_{1}, \dots ,\bm{c}_{T}$. 
The model is trained to identify masked time-steps within a set of distractors. 
Once the model is pre-trained, we fine-tune it with Connectionist Temporal Classification \cite{graves2006connectionist}.

\section{Results}
\begin{table}
\centering
\begin{tabular}{ l | c c | c c }
 & \multicolumn{2}{c}{scratch}\textbf{} & \multicolumn{2}{c}{feat. pretrained}\\
  & dev-o & dev-c & dev-o & dev-c \\ \hline
 pretrained real data & 19.3 & 13.5 & 19.8 & 13.9 \\  
 random initialization & 98.3 & 98.8 & 68.8 & 57.9 \\  
 VCTK  & 38.7 & 24.0 & 31.8 & 20.1  \\  
 LJSpeech & 47.1 & 28.9  & 35.5 & 22.7 \\  
\end{tabular}
\caption{Librispeech dev-clean (dev-c) and dev-other (dev-o) results on systems pre-trained with data from two different synthesizers. Both synthesizers use one only speaker. We provide a topline where the system is pre-trained on the 960 hours of Librispeech (pretrained real data), and a baseline that is not pre-trained (random initialization). We also present results with a pre-trained feature extractor (feat. pretrained).\label{tab:initial_comparission}}
    \vspace*{-\baselineskip}
\end{table}

\subsection{Pre-training with Synthetic Speech}

We start by analyzing if we can pre-train models on synthetic speech. To do so, we pre-train our model using data generated by our two synthesizers. Table \ref{tab:initial_comparission} shows these results and compares them to a non-pre-trained model (\textit{random initialization}) and a model pre-trained on the 960 hours of natural Librispeech data. As expected, all models pre-trained on synthetic speech perform better than the baseline, and underperform the system pre-trained on natural speech. We hypothesize that the gap with the topline model (\textit{pretrained real data}) is due to the small imperfections in the human speech signal, that synthesizers cannot reproduce. Interestingly, we observe that although LJspeech's synthesis quality is higher \footnote{Note that LJSpeech's mean opinion score is considerably higher than VCTK's (see Table 1, and Table 3 in Wang \textit{et al.}\cite{wang-etal-2021-fairseq})}, this improvement is not present in the overall WER performance. We hypothesize that this is due lack of speaker diversity.

\subsubsection{Effect of Speaker Variation}
\begin{table}
\centering
\begin{tabular}{ l | c c | c c }
Speakers & \multicolumn{2}{c}{scratch}\textbf{} & \multicolumn{2}{c}{feat. pretrained}\\
  & dev-other & dev-clean & dev-other & dev-clean \\ \hline
 1 & 54.1 & 39.4 & 37.1 & 22.8 \\  
5 & 48.3 & 33.8 & 37.0 & 23.1\\
10  & 46.6 & 31.2  & 37.0 & 22.6\\  
50 & 40.9 &  26.3 & 33.4 & 19.7\\  
100 & 39.3 & 24.3  &  32.0 & 20.1\\  
\end{tabular}
\caption{Development results on pre-training with different speaker variability. The Speakers column designates the number of speakers used during pre-training. The data for each speaker is distributed homogeneously. We also present results with a pre-trained feature extractor (feat. pretrained).}\label{tab:speaker_variability}
    \vspace*{-\baselineskip}
\end{table}

In the previous section, we hypothesized that speaker diversity in the pre-training set might affect the performance of the fine-tuned model. Now we ask \textit{how important is to have speaker diversity in the pre-training set?} To do so, using VCTK, we synthesize sets with different number of speakers, leaving the remaining parameters of the dataset fixed (\eg\  content, utterances, length). We assign the same number of utterances to each speaker. In Table~\ref{tab:speaker_variability}, we show the results of this experiment. We observe that speaker diversity is important -- although it plateaus after 50 speakers. Note that even though VCTK's synthesis quality is lower, it outperforms the model pre-trained on LJSpeech. This observation stresses the importance of speaker variation in the pre-training set -- this conclusion is consistent with concurrent work \cite{berrebbi2022more}.

\subsection{Semantic Domain}

Next, we investigate the effects of modifying the semantic content of the pre-training set. To do so, we synthesize the transcriptions of Librispeech, Switchboard \cite{godfrey1992switchboard}, and Switchboard+Fisher \cite{cieri2004fisher}, which have a lexicon of 206K and 400K words, respectively. The semantic domain of both datasets is considerably different -- Switchboard is composed for telephone conversations and Librispeech for read speech audiobooks. For this experiment, we use VCTK with only one speaker to not include other domain factors in the the pipeline. To disentangle phonetic and acoustic contributions, we pre-train our model using a frozen feature extractor from a fully-trained model with 960 hours of Librispeech set. Results (Table \ref{tab:linguistic_variability}) show that linguistic mismatch does not have a strong effect. Interestingly, in Hsu \textit{et al.} \cite{hsu2021robust}, when Switchboard is used as a pre-training set, they observe a more pronounced performance degradation (96.2\% relative) than in our experiment (7\% relative). Both results indicate that a big part of the performance gap is due to the acoustic rather than lexical mismatch. Finally, we observe that by adding the Fisher subset (\ie\ increasing lexicon and hours of data), the model does not recover any performance. This result further reveals that the semantic factor is trivial during pre-training.

\begin{table}

\centering
\begin{tabular}{l c c c c }
corpus & dev-other & dev-clean \\ \hline
librispeech &  37.1 & 22.8 \\  
switchboard  & 39.7 & 25.3\\
switchboard + Fisher   & 39.7 & 25.0  \\  
\end{tabular}
\caption{Results on Semantic domain. We synthesize the transcriptions of Librispeech, Switchboard, and Switchboard+Fisher using VCTK (only one speaker), pre- train models on this data, fine-tune using LL-10h and measure performance on Librispeech dev-other/dev-clean. All models use a pre-trained feature extractor.}\label{tab:linguistic_variability}
\end{table}

\subsection{Syntactic Domain}
\begin{table}
\begin{tabular}{ l |  c | c c }
  & type & dev-other & dev-clean \\ \hline
    baseline & synth & 35.5 & 22.7 \\
    word shuffled  & synth  & 36.5 & 23.0\\ \hline
      baseline & mod & 19.8 & 13.9 \\ 
    word shuffled & mod & 23.9 & 16.5 \\  
    word random shuffled & mod & 30.5 & 22.7\\  
\end{tabular}
\caption{Results on shuffling the order of words either by modifying the original audio (mod), or synthesizing a sequence of shuffled words (synth). \textit{word random shuffled} uses the same number of segments and duration as \textit{word shuffled} but with a random start time. Synth, and mod baselines are systems pre-trained on the unmodified synthesized and audio pre-training sets.}\label{tab:grammar_variation}
     \vspace*{-\baselineskip}
\end{table}

We can broadly define the syntactic domain as the structure used to combine words and construct sentences. A syntactic mismatch between pre-training and fine-tuning is a common phenomenon. For example, one could pre-train on YouTube speech, which usually contains more syntactic variations, and then fine-tune the same model on a domain with a clear syntactic structure (\eg\ audiobooks speech). To explore the effects of syntactic mismatch, we create a pre-training set where we randomly shuffle the word order of each utterance and synthesize this data with the LJSpeech model (\textsection\ref{sec:datasetcreation}). This experiment represents a strong mismatch in the syntactic domain, where the grammar structure during pre-training is nonexistent. Similar to the previous section, to isolate the contribution of syntactic mismatch and remove feature extraction effects, we only consider models with a pre-trained feature extractor. Table~\ref{tab:grammar_variation} shows the results of this experiment on synthetic speech and modified audio. In the synthetic speech experiments, we observe a slight degradation in performance (0.1\%, and 3\% relative WER for dev-clean, and dev-other respectively). This result indicates that syntactic mismatch has minor effect during pre-training. 

\begin{figure}

\begin{tikzpicture}
\node[inner sep=0pt] (russell) at (0,0)
    {\includegraphics[width=\linewidth]{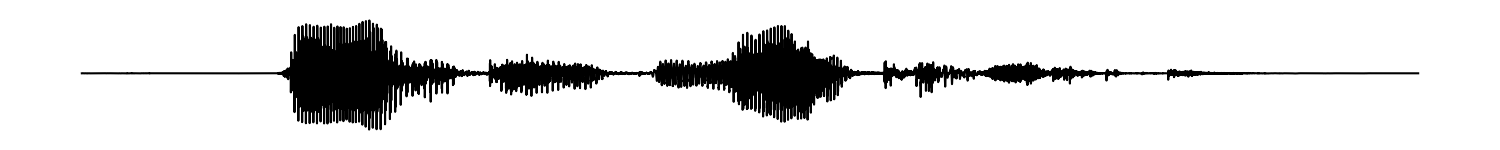}};
\node[inner sep=0pt] (whitehead) at (0,-1.7)
    {\includegraphics[width=\linewidth]{figs/waveform.png}};

\node[rectangle,
        fill={black!60!green},
        opacity=0.4,
        minimum width = 1.1cm,
        minimum height = 0.8cm,
        outer sep=0] (firstword) at (-2,0) {};

\node[rectangle, 
    fill={black!60!red},
    opacity=0.4,
    minimum width = 0.9cm, 
    minimum height = 0.8cm, 
    anchor=north west,
    outer sep=0] (secondword) at (firstword.north east) {};
    
\node[rectangle, 
    fill={rgb:red,4;green,2;yellow,1},
    opacity=0.4,
    minimum width = 1.1cm, 
    minimum height = 0.8cm,         
    anchor=north west,
    outer sep=0] (thirdword) at (secondword.north east) {};
    
\node[rectangle, 
    fill={rgb:red,1;green,2;blue,5},
    opacity=0.4,
    minimum width = 2cm, 
    minimum height = 0.8cm, 
    anchor=north west,
    outer sep=0] (fourthword) at (thirdword.north east) {};

\node[align=left,black,anchor=south] at (firstword.north) {I};

\node[align=left,black,anchor=south] at (secondword.north) {GIVE};

\node[align=left,black,anchor=south] at (thirdword.north) {MY};

\node[align=left,black,anchor=south] at (fourthword.north) {CONSENT};

\node[rectangle,
        fill={rgb:red,1;green,2;blue,5},
        opacity=0.4,
        minimum width = 2cm,
        minimum height = 0.8cm,
        outer sep=0] (firstwordran) at (-1.55,-1.7) {};

\node[rectangle, 
    fill={black!60!green},
        opacity=0.4,
    minimum width = 1.1cm, 
    minimum height = 0.8cm, 
    anchor=north west,
    outer sep=0] (secondwordran) at (firstwordran.north east) {};

\node[rectangle, 
    fill={black!60!red},
    opacity=0.4,    
    minimum width = 0.9cm, 
    minimum height = 0.8cm, 
    anchor=north west,
    outer sep=0] (thirdwordran) at (secondwordran.north east) {};

\node[rectangle, 
    fill={rgb:red,4;green,2;yellow,1},
    opacity=0.4,
    minimum width = 1.1cm, 
    minimum height = 0.8cm, 
    anchor=north west,
    outer sep=0] (fourthwordran) at (thirdwordran.north east) {};

\draw[<->,thick, draw={black!60!green}] (firstword.south) to[out=-90,in=90] (secondwordran.north)
    node[midway] {};

\draw[<->,thick, draw={rgb:red,1;green,2;blue,5}] (fourthword.south) to[out=-90,in=90] (firstwordran.north)
    node[midway] {};

\draw[<->,thick,draw={rgb:red,4;green,2;yellow,1}] (thirdword.south) to[out=-90,in=90] (fourthwordran.north)
   node[midway] {};

\draw[<->,thick, draw={black!60!red}] (secondword.south) to[out=-90,in=90] (thirdwordran.north)
   node[midway] {};

\end{tikzpicture}
    \caption{Illustration of \textit{wrd rnd} (bottom) using ground truth segments (up) shuffled. 
    }    \label{fig:wrdrndspan}
\end{figure}

Because synthetic speech suffer from artifacts, we repeat the word random shuffle experiment using the original audio files (\ie cutting, reordering, and concatenating back again) -- instead of using synthesis. Note that in this case, some artifacts may appear in the discontinuities where audio chunks are concatenated. As a random baseline, we prepare a set that uses the same number of segments and duration but we randomize the start time of the segmentation process (see Figure ~\ref{fig:wrdrndspan}). Note that this experiment perturbs the inner structure of words/morphemes (\ie\ phonotactics).  Table~\ref{tab:grammar_variation} (\textit{mod} columns) presents these results. Compared to the synthesized experiments (\textit{synth} row), we observe higher performance degradation when randomizing the word order on the audio signal. We hypothesize that this performance difference is due to the discontinuities added during audio modification. Results show further degradation in performance when using random segmentation (\textit{word random shuffled} row). This result suggests that a matching inner word structure -- \ie\ phonotactics -- between the pre-training and fine-tuning sets is important. We further explore this observation next.

\begin{table}
\centering
\begin{tabular}{ l |  c | c | c  }
  & type & dev-other & dev-clean \\ \hline
     baseline & synth & 35.5 & 22.7 \\
         phone shuffled  & synth & 46.6 & 30.9\\  
        \hline
             baseline & mod & 19.8 & 13.9 \\
  phone shuffled  & mod  & 40.6 & 30.2 \\  
 phone random shuffled  & mod & 39.5 &  29.2\\  
  no pre-training & mod & 68.8 & 57.9 \\  
\end{tabular}
\caption{Results on phone shuffling during pre-training on synthesized (synth), and the original audio (mod) using a pre-trained feature extractor. We compare these results with a non-pretrained baselin (\textit{no pre-training}), and with system pre-trained on the 960 hour set from Librispeech (\textit{baseline}).}\label{tab:data_modification_phn}
     \vspace*{-\baselineskip}
\end{table}

\subsection{Phonetic Domain}

Phonotactics define the rules languages follow to create possible phone sequences (\ie\ morphemes). To study the effect of a phonotactic mismatch, we generate a pre-training set where phones are randomly shuffled using the LJSpeech synthesizer and segmenting and reordering phones in the audio signal. To eliminate acoustic feature learning from our analysis, we restrict this set of experiments to systems with a pre-trained feature extractor. Table~\ref{tab:data_modification_phn} presents results on pre-training on synthesized shuffled phone sequences and audio modified experiments. In general (\ie in audio modification and synthesis experiments), we observe a larger degradation on phone than in word shuffling. These results further suggest that phonetic consistency is more important than syntactic consistency during pre-training. We also observe that audio modification experiments suffer a larger degradation than the synthesized ones. Again, similar to the word-level experiment, this difference in performance suggests that part of the degradation is due to the discontinuities added during the audio modification process.




\begin{table}
\centering
\begin{tabular}{ l | c c | c c }
 & \multicolumn{2}{c}{scratch}\textbf{} & \multicolumn{2}{c}{feat. pretrained}\\
words & dev-other & dev-clean & dev-other & dev-clean \\ 
\hline
 2 & 50.8 &  34.2 & 42.2 &  26.2 \\  
 3 & 48.9 & 32.2 & 36.2 & 21.6 \\  
 5 & 49.9 & 32.4 & 36.4 & 20.5 \\
 \hline
no seg. & 47.1 & 28.9  & 35.5 & 20.2 \\  
\end{tabular}
\caption{Effect of prosody variation of speech synthesized with the LJSpeech model. The first column determines the number of of words synthesized in isolation before concatenation. Feat. pretrained indicates that we used features from a fully pretrained.}\label{tab:prosody_varibility} 
    \vspace*{-\baselineskip}
\end{table}\unskip

Prosody is a rhythm, stress, and intonation at the utterance level. Here we ask how important is prosody during pre-training. To answer this question we use the LJspeech synthesizer, which can generate better prosody due to being trained with longer sentences. To remove prosody, we synthesize groups of words in isolation so that the synthesizer does not model their phonetic interaction. Table~\ref{tab:prosody_varibility} shows results synthesizing groups of a different number of words in isolation using a pre-trained feature extractor, and a model trained from scratch.  We observe that the larger the prosody span the better. This suggests that prosody is an important factor to consider during pre-training.

\begin{table}
\centering
\begin{tabular}{ l  | c c | c c }
&\multicolumn{2}{c}{scratch}\textbf{} & \multicolumn{2}{c}{feat. pretrained}\\ 
    method & dev-o & dev-c & dev-o & dev-c \\ \hline
    no pre-training  & 98.3 & 98.8 & 68.8 & 57.9 \\
    white noise & -  & - & - & - \\  
    word noise seq.  & - & - & 71.2 & 87.9 \\  
    phone tone seq.  & - & - & 59.4 & 47.6 \\  
\end{tabular}
\caption{Results on pre-training with the random tone and noise sequences of phone length (\textit{phone tone seq}) and word length (\textit{word noise seq.}). We compare these systems to a system trained on white noise (\textit{white noise}), and a system without pre-training (\textit{no pre-training}). The symbol  `-` designates that the model did not convergence. Feat. pretrained indicates that we used features from a fully pre-trained.}\label{tab:synth_lan}
\vspace*{-\baselineskip}
\end{table}

Previously, we observe that pseudo-randomized signals (\ie\ phonotactically inconsistent phone sequences) can pre-train the Transformer module of wav2vec 2.0 to a certain extent. Now, we test if the Transformer module can learn from a synthetic language composed for a synthetic phone set. This synthetic phone set, consists of 44 sounds -- white, brown, pink, blue, violate noises, and tones sampled from 200Hz to 900Hz (approximate range of the first formants). Each sound has a duration of 90ms (average phone duration in Librispeech) with a sampled $\pm$30ms interval and sampled volume. We construct a lexicon by randomly sampling character lengths and sequences of phonemes (from our 44-phoneme set). We sample word lengths and words to create sentences. All distributions that we sample from are uniform. Table \ref{tab:synth_lan} in \textit{phone tone seq.} row shows the results of models pre-trained on the synthetic language. First, we observe that the model requires a pre-trained feature extractor to use our synthetic data -- when the feature extractor is pre-trained from scratch, the models diverge. Surprisingly, although the synthetic language does not share any acoustic similarity, syntactic, phonotactic, or lexical, it still outperforms a non-pre-trained baseline. This result is consistent with previous work on NLP that successfully pre-train Transformer models with out-of-domain \cite{papadimitriou-jurafsky-2020-learning} or pseudo-random \cite{krishna2021does} data. As random baselines, we experiment with pre-training the model on fewer structure data. Concretely, we experiment using Gaussian noise (\textit{white noise}), and sequences of tones and noises similar to the synthetic languages, but instead of using phone units, we randomly sample with a duration of 300ms with a $\pm$30ms sampled variation (\textit{word noise seq.}) -- the average duration of a word. Results show that when pre-training on these sets the system either underperforms the baseline or can not be trained.


\subsection{Large Synthetic Set vs. Small Real Set}

\begin{table}

\begin{tabular}{ l | c | c c | c c }
 &  &\multicolumn{2}{c}{scratch}\textbf{} & \multicolumn{2}{c}{feat. pretrained}\\ 
    source & data & dev-o & dev-c & dev-o & dev-c \\ \hline
    synth (VCTK) & LS & 38.7 & 24.0 & 31.8 & 20.1  \\  
    human & VCTK & 49.7 & 36.2 & 38.2 & 26.1 \\ \hline
    synth (LJ) & LS & 47.1 & 28.9  & 35.5 & 20.2  \\  
    human & LJ &49.8 & 33.1 & 36.5 & 22.7 \\ 
\end{tabular}
\caption{Comparison of models pretrained on synthesized Librispeech (\textit{synth}) and real speech data used to train the synthesizer (\textit{human}). The Source column specify how the data has been generated. The data column which pre-training set is used. For instance the first row a synthesizer trained on VCTK generates the Librispeech data set. The VCTK model is pretrained with 109 speakers. Feat. pretrained indicates that we used features from a fully pre-trained.}\label{tab:synth_vs_real}
     \vspace*{-\baselineskip}
\end{table}

Ultimately, as a supplementary experiment, we compare the performance of pre-training on a large synthesized pre-training set and the real data used to train the synthesizer. Concretely, we synthesize all Librispeech using VCTK or LJspeech (a total 960 hours of synthesized speech with the available speakers in each synthesizer), and compare it with a model pre-trained on the training set for each syntheszer (24, and 44 hours respectively). Table~\ref{tab:synth_vs_real} show the results of this comparison. We observe that the large-synthesized sets, outperform the smaller ones with real speech -- specially for VCTK. This result suggests that when only few data is available for pre-training, one could use a synthesizer to augment the amount of data to improve results.

\section{Conclusions and future work}
We present a controlled study to understand how individual domain factors affect self-supervised speech representation learning. With the help of speech synthesis and audio modification we alter specific domain dimensions of the pre-training set and measure performance on ASR. We observe that phonetic domain is much more important than altering the syntactic and semantic domain. Initial experiments indicate that wav2vec2.0 can be pre-trained on sequences of synthetic noise sequences that does not share any phonetic, semantic nor syntactic similarity with the target data.

In terms of future work, we plan to experiment with other architectures such as HuBERT \cite{hsu2021hubert}, and confirm our results with human generated data.
\bibliographystyle{IEEEtran}
\bibliography{mybib}

\end{document}